\documentclass[runningheads]{waica}
\usepackage[T1]{fontenc}
\usepackage{amsmath}
\usepackage{graphicx}
\usepackage{amsfonts}
\usepackage{multirow}
\usepackage{hyperref}
\usepackage{color}

\usepackage{booktabs}
\usepackage[table, xcdraw]{xcolor}

\begin{document}
\title{XrayClaw: Cooperative-Competitive Multi-Agent Alignment for Trustworthy
Chest X-ray Diagnosis}
\titlerunning{XrayClaw}



\author{Shawn Young \and
Lijian Xu\thanks{Corresponding author.}
}

\institute{Shenzhen University of Advanced Technology, Shenzhen, China}

\institute{Shenzhen University of Advanced Technology, Shenzhen, China
}

\maketitle 
\begin{abstract}
Chest X-ray (CXR) interpretation is a fundamental yet complex clinical task that increasingly relies on artificial intelligence for automation. However, traditional monolithic models often lack the nuanced reasoning required for trustworthy diagnosis, frequently leading to logical inconsistencies and diagnostic hallucinations. While multi-agent systems offer a potential solution by simulating collaborative consultations, existing frameworks remain susceptible to consensus-based errors when instantiated by a single underlying model. 
This paper introduces XrayClaw, a novel framework that operationalizes multi-agent alignment through a sophisticated cooperative-competitive architecture. XrayClaw integrates four specialized cooperative agents to simulate a systematic clinical workflow, alongside a competitive agent that serves as an independent auditor. To reconcile these distinct diagnostic pathways, we propose Competitive Preference Optimization, a learning objective that penalizes illogical reasoning by enforcing mutual verification between analytical and holistic interpretations. Extensive empirical evaluations on the MS-CXR-T, MIMIC-CXR, and CheXbench benchmarks demonstrate that XrayClaw achieves state-of-the-art performance in diagnostic accuracy, clinical reasoning fidelity, and zero-shot domain generalization. Our results indicate that XrayClaw effectively mitigates cumulative hallucinations and enhances the overall reliability of automated CXR diagnosis, establishing a new paradigm for trustworthy medical imaging analysis.

\keywords{Multi-Agent Systems \and Trustworthy Diagnosis \and Chest X-ray.}
\end{abstract}

\section{Introduction}

Chest X-ray (CXR) imaging remains the most frequently utilized diagnostic tool in clinical practice due to its cost-effectiveness and accessibility in various healthcare settings \cite{xu2024foundation}. Recent advancements \cite{yang2025walking,young2025fewer,yangone,yang2026scalar} in artificial intelligence have significantly transformed the landscape of medical imaging analysis \cite{feng2026efficient,chen2026tc,yang2024segmentation}, with deep learning models demonstrating remarkable proficiency in tasks such as abnormality classification and lesion localization. However, the inherent complexity of CXR interpretation, which often necessitates nuanced reasoning and the integration of multi-modal information \cite{yang2026scalar,he2026model,yu2026drift,yang2023t}, has catalyzed a paradigm shift toward more sophisticated architectural frameworks \cite{xu2023learning,xu2024medvilam}. Within this context, multi-agent system (MAS) have emerged as a promising frontier, offering the capacity to model the collaborative nature of human radiological consultations \cite{fallahpour2025medrax,wei2024medco,yang2022local}. By leveraging the collective intelligence of multiple specialized agents, MAS holds immense potential to enhance the accuracy and trustworthiness of automated CXR diagnosis, effectively addressing the limitations of traditional monolithic models in handling diverse and complex clinical scenarios.

Despite the advantages of MAS in medical diagnoise, existing frameworks encounter significant challenges regarding diagnostic trustworthiness and clinical alignment \cite{chen2024chexagent,yangadapting,yang2025resilient}. When these systems rely on a single underlying model to instantiate multiple functional roles, they are particularly susceptible to internal consistency errors and cumulative hallucinations. Specifically, current methods often fail to establish a robust verification mechanism that aligns the collaborative reasoning of agents with established medical gold standards \cite{yang2025learning,zhou2025mam}. This lack of rigorous alignment frequently results in consensus-based errors, where agents converge on a diagnosis that lacks clinical validity, thereby undermining the overall reliability of the diagnostic output. Furthermore, maintaining critical objectivity within a single-model-based MAS remains difficult, as the absence of diverse architectural perspectives can lead to overconfidence in erroneous predictions. Consequently, bridging the gap between multi-agent interaction and trustworthy clinical decision-making is an unresolved challenge that limits the practical deployment of MAS in CXR diagnosis.

To address these challenges in diagnostic trustworthiness and clinical alignment, we introduce a cooperative-competitive multi-agent framework designed to emulate the rigorous scrutiny of clinical peer reviews. In this architecture, cooperation among agents facilitates the exchange of diverse feature representations, thereby enhancing the coverage of clinical information and ensuring a comprehensive analysis of the radiographic evidence. Simultaneously, a competitive mechanism is integrated through the preference optimization, wherein agents are incentivized to identify logical vulnerabilities and inconsistencies in the propositions of their counterparts. This adversarial interaction serves as a critical self-correction layer, effectively mitigating the risk of cumulative hallucinations and strengthening the robustness of the diagnostic process. By balancing these dual dynamics, the system achieves a higher degree of alignment with medical logic, ensuring that the final diagnostic output is both accurate and justifiable.

In summary, our paper mainly makes the following contributions:

\begin{itemize}
    \item We propose XrayClaw, a novel framework that operationalizes multi-agent alignment through a sophisticated cooperative-competitive architecture. By orchestrating four specialized cooperative agents to simulate the systematic clinical workflow and incorporating an independent auditor, the framework emulates the rigorous scrutiny of clinical peer reviews to significantly enhance diagnostic traceability and reliability.

    \item We introduce Competitive Preference Optimization (ComPO), an innovative alignment objective designed to facilitate preference comparisons between the analytical reasoning of the cooperative pipeline and the holistic predictions of the omni-radiologist. ComPO effectively identifies and penalizes logical hallucinations by enforcing mutual verification between distinct diagnostic pathways, thereby incentivizing the convergence of agent outputs toward established clinical gold standards.

    \item We conduct extensive empirical evaluations across three authoritative benchmarks. The results demonstrate that XrayClaw achieves superior performance in diagnostic accuracy on MS-CXR-T, clinical reasoning fidelity on MIMIC-CXR, and domain generalization on CheXbench.
\end{itemize}

\section{Methodology}

\begin{figure}[ht]
    \centering
    \includegraphics[width=1\linewidth]{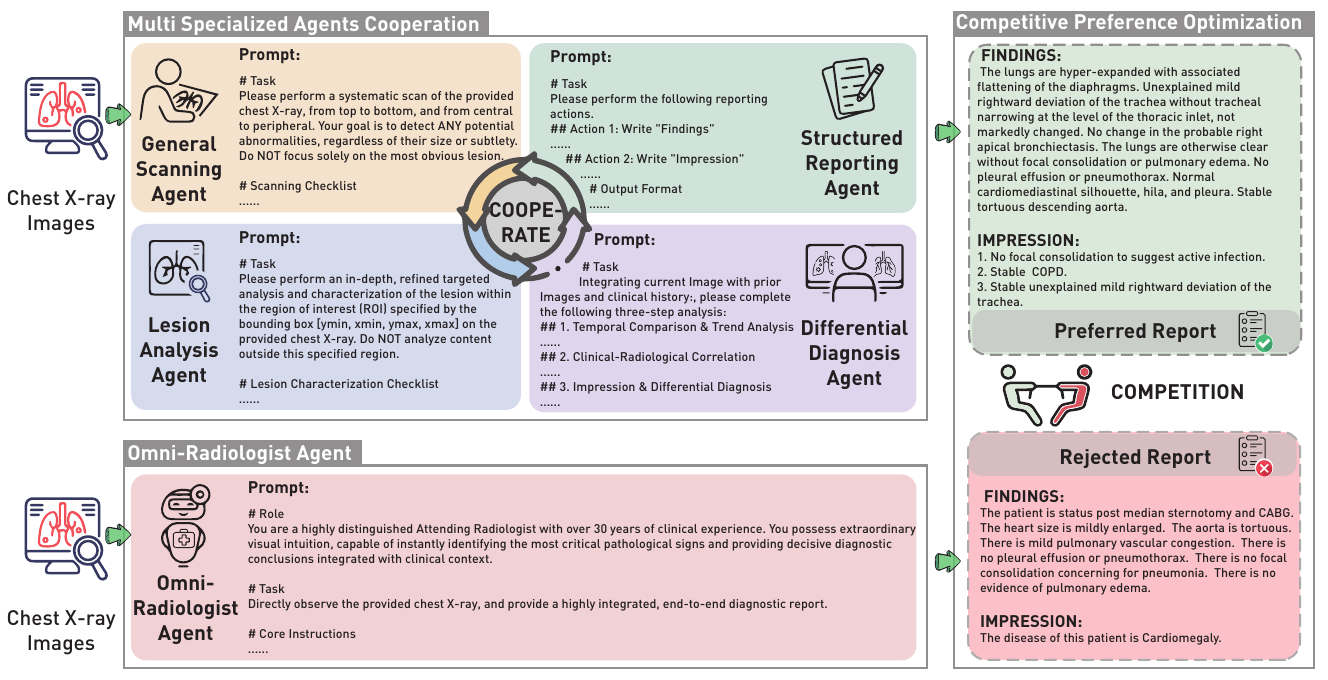}
    \caption{The overarching architecture of XrayClaw. The framework orchestrates
    a cooperative-competitive alignment process for trustworthy Chest X-ray
    diagnosis. The Multi Specialized Agents Cooperation pipeline (top)
    decomposes the diagnostic task into four sequential stages: systematic scanning,
    targeted lesion analysis, differential reasoning, and structured report
    synthesis. In parallel, the Omni-Radiologist Agent (bottom) generates a
    holistic, end-to-end diagnostic output. Through Competitive Preference
    Optimization, the framework performs a preference comparison between
    the two pathways, incentivizing the model to favor the rigorous, evidence-based
    reasoning of the cooperative pipeline over potential hallucinations in the
    end-to-end output, thereby ensuring clinical trustworthiness.}
    \label{fig:workflow}
\end{figure}

\subsection{Cooperative Multi-Agent Diagnostic}

The design of the cooperative multi-agent diagnostic pipeline in XrayClaw is motivated by the observation that traditional LLM frameworks often treat CXR diagnosis as a monolithic classification or generation task. Such approaches frequently overlook the rigorous observation-analysis-inference logic that defines the professional workflow of a radiologist. To address this limitation, the core philosophy of XrayClaw involves the deconstruction of complex medical diagnosis into a sequence of atomic clinical sub-tasks. By adopting this divide-and-conquer paradigm, the framework enhances the traceability and interpretability of the diagnostic process, thereby establishing a foundation for clinical trustworthiness that is often absent in black-box monolithic models.

First, the transition from global context to local scrutiny is operationalized through a hierarchical coordination between specialized agents. A singular general scanning agent is first deployed to perform a comprehensive primary screening of the radiographic image. The objective of this agent is to establish spatial coordinates and identify fundamental anatomical hallmarks, such as the symmetry of lung fields and the sharpness of costophrenic angles, resulting in a set of global abnormality leads. Subsequently, for every specific lead identified during the global scan, a dedicated instance of the lesion analysis agent is invoked to perform a fine-grained evaluation. This targeted analysis focuses on the extraction of microscopic features, including the morphology, margins, and density of the suspected lesion. This evidence flow, moving from localization to characterization, ensures that every diagnostic conclusion is grounded in specific radiographic evidence.

Subsequently, the synthesis of these characterizations into a final clinical judgment is managed by the subsequent reasoning agents to achieve a logical closed loop. The differential diagnosis agent serves as the cognitive core of the system, synthesizing the multiple diagnostic results provided by the various lesion analysis agents. Beyond mere aggregation, this agent performs a verification of the findings and integrates external clinical context, such as patient history and temporal trends in prior imaging, to execute exclusionary reasoning. This process is critical for resolving the challenge of similar shadows, where distinct pathologies may manifest with similar radiographic appearances. Finally, the structured reporting agent acts as the standardized interface of the system, transforming non-structured reasoning into a formal report that adheres to clinical protocols, such as the Findings and Impression sections of standard radiological reports. This architectural step ensures the alignment of the system output with medical gold standards.

The internal consistency and coherence of the diagnostic process are maintained through a robust communication protocol and a traceable chain of evidence. XrayClaw utilizes a sequential prompting mechanism wherein the output of each agent is encapsulated and transmitted as contextual input to the next stage of the pipeline. This collaborative interaction is designed not as a simple concatenation of text, but as a cumulative process of information gain. By maintaining this continuous chain of evidence, the framework ensures that every statement in the final report is supported by a verifiable sequence of observations and logical deductions. This rigorous maintenance of the evidence chain significantly mitigates the risk of logical inconsistencies and reinforces the overall reliability of the multi-agent system.

\subsection{Orchestrating Competitive Auditing for CXR Diagnosis}

Motivated by the inherent risks of sequential processing within the cooperative multi-agent diagnostic pipeline, we introduce the \textbf{Senior Attending Radiologist} agent as a competitive counterpart. While the cooperative pipeline is designed for meticulous evidence extraction, it remains susceptible to the propagation of errors and local optima, where an initial misinterpretation by an antecedent agent leads to a cascade of logical inconsistencies. In a clinical environment, this dynamic corresponds to the hierarchical relationship between a resident, who performs systematic checklist-based analysis, and a senior attending physician, who provides a holistic final audit based on decades of accumulated experience. By incorporating this competitive agent as a high-level auditor, the framework facilitates a direct confrontation between the analytical reasoning of the pipeline and an end-to-end expert interpretation, thereby identifying latent logical conflicts.

The senior attending radiologist is instantiated through a specialized prompting agent that simulates the expertise of a practitioner with thirty years of clinical experience. This agent is characterized by its end-to-end diagnostic capability, bypassing the intermediate stages of scanning and localized analysis to derive clinical impressions directly from the image pixels. The input for this agent comprises the radiographic image alongside highly integrated clinical context instructions, which are designed to emulate the intuition of an expert when presented with complex pathological manifestations. Unlike the atomized tasks of the cooperative pipeline, this holistic approach prioritizes the global synthesis of visual and clinical information to ensure a comprehensive diagnostic perspective.

The interaction between the analytical report generated by the cooperative pipeline and the intuitive report provided by the senior attending agent establishes a parallel execution mechanism. This duality serves as a conflict trigger; whenever a significant discrepancy arises between the two outputs, the resulting inconsistency is captured as a critical signal for model refinement. This confrontation between analytical reasoning and holistic intuition generates high-quality preference pairs, which serve as the primary drivers for the subsequent ComPO. Consequently, this expert-led audit mechanism ensures the trustworthiness of the final diagnostic output by reconciling divergent clinical narratives through rigorous preference alignment.

\subsection{Competitive Preference Optimization}

The core of the XrayClaw framework lies in the ComPO, which serves as the bridge between multi-agent interaction and structural model alignment. This mechanism is theoretically grounded in the dual-process theory of cognitive science, where the analytical reasoning of the cooperative pipeline and the holistic intuition of the senior attending agent are integrated into a unified optimization objective. By utilizing the divergent outputs from these two distinct diagnostic pathways, ComPO operationalizes an adversarial alignment process that incentivizes the model to distinguish between rigorous evidence-based logic and erroneous clinical hallucinations. This strategic interaction ensures that the final diagnostic policy remains anchored in clinical validity.

The construction of preference pairs within ComPO is derived from the confrontation between the analytical trajectory $y_{\text{coop}}$ and the intuitive trajectory $y_{\text{comp}}$. These diagnostic outputs serve as the source for the preference dataset $\mathcal{D}= \{(x, y_{w}, y_{l})\}$, where $x$ denotes the input CXR image. A trajectory is designated as the preferred sample $y_{w}$ if it exhibits superior logical consistency and aligns with the clinical ground truth, while the alternative is designated as the rejected sample $y_{l}$. In scenarios where the two diagnostic pathways yield conflicting conclusions, the ground truth acts as the ultimate oracle to adjudicate the preference. This automated alignment mechanism allows XrayClaw to perform high-fidelity self-correction without the need for exhaustive manual labeling, effectively capturing the nuanced expertise required for trustworthy diagnosis.

To formalize this alignment, we consider a reward function $r(x, y)$ that evaluates the clinical fidelity of a diagnostic report $y$. Following the theoretical framework of Direct Preference Optimization \cite{rafailov2023direct}, the objective is to maximize the expected reward while maintaining a constraint on the Kullback-Leibler (KL) divergence between the learning policy $\pi_{\theta}$ and the reference policy $\pi_{\text{ref}}$:

\begin{equation}
    \max_{\pi_{\theta}}\mathbb{E}_{x \sim \mathcal{D}, y \sim \pi_{\theta}(y|x)}
    [r(x, y)] - \beta \mathbb{D}_{\text{KL}}(\pi_{\theta}(y|x) || \pi_{\text{ref}}
    (y|x))
\end{equation}

where $\beta$ is a regularization parameter that controls the deviation from the base reference policy. The analytical solution for the optimal policy satisfies:
\begin{equation}
    \pi_{\theta}(y|x) \propto \pi_{\text{ref}}(y|x) \exp ( r(x, y) / \beta )
\end{equation}

By reparameterizing the reward function in terms of policy log-probabilities and adopting the Bradley-Terry model, the probability of preferring $y_{w}$ over $y_{l}$ is defined by the sigmoid of the reward difference. The resulting ComPO loss function is formulated as follows:

\begin{equation}
    \mathcal{L}_{\text{ComPO}}(\pi_{\theta}; \pi_{\text{ref}}) = - \mathbb{E}
    _{(x, y{w}, y_{l}) \sim \mathcal{D}}\left[ \log \sigma \left( \beta \log
    \frac{\pi_{\theta}(y_{w}|x)}{\pi_{\text{ref}}(y_{w}|x)}- \beta \log \frac{\pi_{\theta}(y_{l}|x)}{\pi_{\text{ref}}(y_{l}|x)}
    \right) \right]
\end{equation}


In this context, ComPO facilitates a fundamental transition from the superficial mimicry of radiological language to the deep emulation of clinical logic. By minimizing the log-likelihood margin between preferred and rejected diagnostic paths, the framework effectively penalizes outputs that lack verification from the multi-agent evidence chain. This dynamic policy reinforcement ensures that the final structured reports are both interpretable and reliable, achieving a stable equilibrium in complex diagnostic environments. Consequently, the integration of cooperative reasoning and competitive auditing through ComPO establishes a new paradigm for trustworthy medical imaging analysis.

\subsection{Implementation Details}

The implementation of XrayClaw is built upon the Qwen3-VL \cite{bai2025qwen3} architecture, which serves as the foundational multi-modal backbone for all specialized agents. During the initial stage, we perform Supervised Fine-Tuning (SFT) on the MIMIC-CXR training set to ensure the model acquires basic radiological terminology and image-text alignment capabilities. For the cooperative-competitive framework, the system prompts are meticulously designed to delineate the functional roles of the four cooperative agents and the senior attending auditor. All radiographic images are processed using the native vision encoder of Qwen3-VL to extract representative visual features suitable for complex diagnostic reasoning.

The training process for \textbf{ComPO} is conducted using the AdamW optimizer with a global batch size of 4. We set the learning rate for the alignment phase to $5 \times 10^{-6}$ with a cosine learning rate scheduler. The reference policy $\pi_{\text{ref}}$ is initialized from the checkpoint of the SFT model. To facilitate the parallel execution of the cooperative pipeline and the competitive agent, we fine-tune our model across $2 \times 2$ NVIDIA A100 (80GB) GPUs.

During the inference phase, the structured reports are generated using a beam search strategy with a beam width of 3 to enhance the coherence of the diagnostic reasoning. The maximum generation length is set to 512 tokens to accommodate detailed findings and clinical impressions. The entire framework is implemented in PyTorch, and the multi-agent communication protocol is managed through a centralized context buffer that records the cumulative evidence chain for subsequent preference ranking.

\section{Experiments}

\subsection{Trustworthy Disease Diagnosis}

To evaluate the robustness of diagnostic decision-making, we conduct a multi-label classification task on the MS-CXR-T benchmark \cite{bannur2023learning}. As demonstrated in Table \ref{tab:decision}, XrayClaw achieves a state-of-the-art average Top-1 accuracy of 77.9\%, significantly outperforming existing MLLMs-based methods and contemporary multi-agent frameworks. Specifically, XrayClaw surpasses the strongest multi-agent baseline, MedRAX \cite{fallahpour2025medrax}, by a margin of 4.8\% in average accuracy. Compared to high-performing monolithic models such as CoCa-CXR \cite{chen2025cocacxrcontrastivecaptionerslearn} and BioViL-T \cite{bannur2023learning}, the proposed framework exhibits a substantial performance leap, which underscores the efficacy of the cooperative-competitive alignment mechanism in handling complex clinical diagnostic scenarios.

\begin{table}[htp]
    \centering
    \caption{\textbf{Comparative performance of multi-label diagnostic
    classification on the MS-CXR-T benchmark
    \cite{bannur2023learning}.} It reports the Top-1 accuracy
    involving five fundamental pulmonary pathologies, comprising
    consolidation (Con.), pleural effusion (PE), pneumonia (Pna.), pneumothorax
    (Pnx.), and edema (Ede.). The highest and second-highest results are denoted
    in red and blue, respectively.}
    \label{tab:decision}
    \setlength{\tabcolsep}{2mm}{ 
        \begin{tabular}{@{}lccccccc@{}}\toprule & Venue & Con. & PE & Pna. & Pnx. & Ede. & Avg. \\ \midrule 
        \multicolumn{8}{l}{\textit{\textbf{MLLMs-Based}}} \\ 
        CheXRelNet \cite{karwande2022chexrelnet} & MICCAI'22 & 47.0 & 47.0 & 47.0 & 36.0 & 49.0 & 45.2 \\ 
        BioViL \cite{boecking2022making} & ECCV'22 & 56.0 & 63.0 & 60.2 & 42.5 & 67.5 & 57.8 \\ 
        CTransformer \cite{bannur2023learning} & CVPR'23 & 44.0 & 61.3 & 45.1 & 31.5 & 65.5 & 49.5 \\ 
        BioViL-T \cite{bannur2023learning} & CVPR'23 & 61.1 & 67.0 & 61.9 & 42.6 & 68.5 & 60.2 \\ 
        Med-ST \cite{yang2024unlocking} & ICML'24 & 60.6 & 67.4 & 58.5 & 65.0 & 54.2 & 61.1 \\ 
        TempA-VLP \cite{10943948} & WACV'25 & 65.2 & 59.4 & {\color[HTML]{FF0000} 73.4} & 43.1 & {\color[HTML]{4472C4} 77.1} & 63.6 \\ 
        CoCa-CXR \cite{chen2025cocacxrcontrastivecaptionerslearn} & MICCAI'25 & 70.4 & 69.6 & 61.4 & 72.8 & 71.8 & 69.2 \\ \midrule 
        \multicolumn{8}{l}{\textit{\textbf{Multi-Agent}}} \\ 
        RadFabric \cite{chen2025radfabric} & Arxiv‘25 & {\color[HTML]{4472C4} 73.7} & 45.0 & {\color[HTML]{4472C4} 73.3} & {\color[HTML]{4472C4} 85.0} & 66.7 & 68.7 \\ 
        MedRAX \cite{fallahpour2025medrax} & ICML'25 & 69.3 & {\color[HTML]{FF0000} 76.5} & 62.7 & 83.6 & 73.2 & {\color[HTML]{4472C4} 73.1} \\ 
        \rowcolor[HTML]{D9D9D9} 
        {\color[HTML]{333333} XrayClaw (Ours)} & {\color[HTML]{333333} } & {\color[HTML]{FF0000} 73.9} & {\color[HTML]{4472C4} 74.9} & {\color[HTML]{333333} 72.7} & {\color[HTML]{FF0000} 88.7} & {\color[HTML]{FF0000} 79.3} & {\color[HTML]{FF0000} 77.9} \\ \bottomrule
    \end{tabular}}
\end{table}

The detailed breakdown across five core pulmonary pathologies further highlights the diagnostic precision of the proposed method. XrayClaw attains the highest accuracy in three categories, including consolidation (73.9\%), pneumothorax (88.7\%), and edema (79.3\%). The performance on pneumothorax is particularly remarkable, achieving an absolute improvement of 3.7\% over the previous best result from RadFabric \cite{chen2025radfabric}. Although MedRAX \cite{fallahpour2025medrax} maintains a slight advantage in identifying pleural effusion, XrayClaw remains highly competitive with the second-best performance of 74.9\%. This balanced and superior performance across diverse pathologies suggests that the integration of the analytical cooperative pipeline and the expert-level auditing agent effectively mitigates the risk of individual agent bias and diagnostic hallucinations.

The quantitative advancements presented in this section substantiate the trustworthiness of the diagnostic process within XrayClaw. Unlike traditional models that often struggle to maintain consistency across varying pathological manifestations, the multi-agent alignment within XrayClaw ensures that every diagnostic conclusion is the result of rigorous internal verification. The consistent gains in both average and individual pathology accuracy indicate that the proposed architecture successfully reconciles granular evidence extraction with holistic clinical intuition. Consequently, these results demonstrate that XrayClaw provides a more reliable and robust solution for automated CXR interpretation, fulfilling the requirements for trustworthy medical artificial intelligence.

\subsection{Faithful Clinical Reasoning}

To evaluate the fidelity of the diagnostic reasoning process, we assess the performance of XrayClaw on the diagnostic report generation task using the MIMIC-CXR benchmark \cite{johnson2019mimic}. As presented in Table \ref{tab:reasoning}, the proposed framework achieves superior results across all evaluated metrics, including BLEU-4 (0.163), ROUGE-L (0.327), METEOR (0.190), and CIDEr (0.368). These quantitative improvements over traditional, MLLMs-based, and multi-agent baselines demonstrate that XrayClaw is capable of generating diagnostic chain-of-thought trajectories with higher quality and coherence. Specifically, XrayClaw outperforms the strongest multi-agent competitor, MedRAX \cite{fallahpour2025medrax}, by a significant margin in terms of METEOR (0.190 vs. 0.157) and CIDEr (0.368 vs. 0.297), which are metrics more closely associated with clinical content retention and semantic alignment.

\begin{table}[htbp]
    \centering
    \caption{\textbf{Comparative performance of diagnostic reasoning on the MIMIC-CXR \cite{johnson2019mimic} benchmark.} It reports various metrics, including the BLEU-4, ROUGE-L, METEOR and CIDEr on diagnostic report generation task, to assess the quality and coherence of the generated chain-of-thought trajectories. The highest and second-highest results are denoted in red and blue, respectively.}
    \label{tab:reasoning}
    \setlength{\tabcolsep}{1mm}{
        \begin{tabular}{@{}lccccc@{}}
        \toprule
        Methods                                                       & Venue      & BLEU-4                      & ROUGE-L                     & METEOR                      & CIDEr                       \\ \midrule
        \multicolumn{6}{l}{\textit{\textbf{Traditional}}}                                                                                                                                                  \\
        GSK \cite{yang2022knowledge}                                  & MIA'22     & 11.5                        & 28.4                        & -                           & 20.3                        \\
        Clinical-BERT   \cite{yan2022clinical}                        & AAAI'22    & 10.6                        & 27.5                        & 14.4                        & 15.1                        \\
        METransformer   \cite{wang2023metransformer}                  & CVPR'23    & 12.4                        & 29.1                        & 15.2                        & {\color[HTML]{4472C4} 36.2} \\
        DCL \cite{li2023dynamic}                                      & CVPR'23    & 10.9                        & 28.4                        & 15.0                        & 28.1                        \\
        STREAM \cite{yang2025spatio}                                  & TMI'25     & 13.3                        & 29.1                        & 16.4                        & 0.0                         \\
        DC-SGC   \cite{wangDiagnosticCaptioningCooperative2025}       & TPAMI'25   & 12.9                        & 29.1                        & 16.4                        & 34.2                        \\ \midrule
        \multicolumn{6}{l}{\textit{\textbf{MLLMs-Based}}}                                                                                                                                                  \\
        R2GenGPT   \cite{wang2023r2gengpt}                            & MetaRad'23 & 12.5                        & 28.5                        & 16.7                        & 24.4                        \\
        PromptMRG   \cite{jin2024promptmrg}                           & AAAI'24    & 11.2                        & 26.8                        & 15.7                        & -                           \\
        BtspLLM   \cite{liu2024bootstrapping}                         & AAAI'24    & 12.8                        & 29.1                        & {\color[HTML]{4472C4} 17.5} & -                           \\
        MambaXray   \cite{wangCXPMRGBenchPretrainingBenchmarking2024} & CVPR‘25    & 13.3                        & 28.9                        & 16.7                        & 24.1                        \\ \midrule
        \multicolumn{6}{l}{\textit{\textbf{Multi-Agent}}}                                                                                                                                                  \\
        MedRAX   \cite{fallahpour2025medrax}                          & ICML'25    & {\color[HTML]{4472C4} 16.0} & {\color[HTML]{4472C4} 32.1} & 15.7                        & 29.7                        \\
        \rowcolor[HTML]{D9D9D9} 
        \multicolumn{2}{l}{\cellcolor[HTML]{D9D9D9}RadClaw   (Ours)}               & {\color[HTML]{FF0000} 16.3} & {\color[HTML]{FF0000} 32.7} & {\color[HTML]{FF0000} 19.0} & {\color[HTML]{FF0000} 36.8} \\ \bottomrule
        \end{tabular}
    }
\end{table}

The comparative analysis reveals that XrayClaw effectively addresses the limitations of monolithic MLLMs and traditional architectures. While advanced MLLMs-based models such as MambaXray \cite{wangCXPMRGBenchPretrainingBenchmarking2024} and BtspLLM \cite{liu2024bootstrapping} provide competitive linguistic fluency, they often lack the structural verification required for complex clinical reasoning. In contrast, the cooperative-competitive architecture of XrayClaw ensures that the generated findings are grounded in evidence extracted through the multi-agent pipeline. The substantial gain in the CIDEr score, which is particularly sensitive to the consensus of domain-specific terminology, highlights the ability of the proposed framework to align generated reports with the professional standards of human radiologists.

Furthermore, the consistency in performance gains across multiple evaluation dimensions underscores the trustworthiness of the reasoning process. The high scores in BLEU-4 and ROUGE-L suggest that the generated reports maintain high linguistic similarity to ground-truth clinical documents, while the peak values in METEOR and CIDEr indicate a superior capture of pathological nuances. This suggests that the ComPO effectively penalizes illogical reasoning paths and reinforces the production of faithful clinical narratives. Consequently, the results in Table \ref{tab:reasoning} provide strong empirical evidence that XrayClaw not only achieves state-of-the-art performance in automated report generation but also significantly enhances the reliability of the underlying diagnostic reasoning.

\subsection{Reliable Domain Generalization}

To evaluate the cross-domain robustness of the diagnostic capabilities, we assess the performance of XrayClaw on the CheXbench benchmark \cite{chen2024chexagent}. This evaluation spans three distinct datasets, Rad-Restruct \cite{pellegrini2023rad}, SLAKE \cite{liu2021slake}, and OpenI \cite{demner2012design}, encompassing both Visual Question Answering and fine-grained reasoning tasks. As detailed in Table \ref{tab:generalization}, XrayClaw achieves an overall accuracy of 70.7\%, surpassing all baselines, including specialized multi-agent systems and large-scale commercial models. Specifically, the framework yields the highest results in the SLAKE (85.6\%) and OpenI reasoning (62.1\%) categories. Notably, the 9.5\% improvement over MedRAX \cite{fallahpour2025medrax} in the reasoning task on OpenI highlights the superior ability of the cooperative-competitive architecture to generalize complex clinical logic across varying domain distributions.

\begin{table}[htbp]
    \centering
    \caption{\textbf{Comparative performance of diagnostic reasoning
    generalization on the CheXbench \cite{chen2024chexagent} benchmark.} It reports
    the Top-1 accuracy involving visual QA and fine-grained reasoning task across
    three dataset, including Rad-Restruct \cite{pellegrini2023rad}, SLAKE \cite{liu2021slake}
    and OpenI \cite{demner2012design}. The highest and second-highest
    results are denoted in red and blue, respectively.}
    \label{tab:generalization}
    \setlength{\tabcolsep}{2mm}
    { \begin{tabular}{@{}lccccc@{}}\toprule & & \multicolumn{2}{c}{Visual QA} & Reasoning & \\ \multirow{-2}{*}{Methods} & \multirow{-2}{*}{Venue} & Rad-Restruct & SLAKE & OpenI & \multirow{-2}{*}{Overall} \\ \midrule \multicolumn{6}{l}{\textit{\textbf{MLLMs-Based}}} \\ LLaVA-Med \cite{li2023llava} & NeurIPS'23 & 34.9 & 55.5 & {\color[HTML]{4472C4} 45.8} & 45.4 \\ GPT-4o \cite{hurst2024gpt} & Commercial & 53.9 & {\color[HTML]{4472C4} 85.4} & 51.1 & 63.5 \\ \midrule \multicolumn{6}{l}{\textit{\textbf{Multi-Agent}}} \\ CheXagent \cite{chen2024chexagent} & AAAI'24 Sym' & 57.1 & 78.1 & 59.0 & 64.7 \\ MedRAX \cite{fallahpour2025medrax} & ICML'25 & {\color[HTML]{FF0000} 68.7} & 82.9 & 52.6 & {\color[HTML]{4472C4} 68.1} \\ \rowcolor[HTML]{D9D9D9} XrayClaw (Ours) & & {\color[HTML]{4472C4} 66.3} & {\color[HTML]{FF0000} 85.6} & {\color[HTML]{FF0000} 62.1} & {\color[HTML]{FF0000} 70.7} \\ \bottomrule\end{tabular} }
\end{table}

The performance of XrayClaw across heterogeneous datasets underscores the reliability of the proposed multi-agent alignment mechanism. Unlike LLaVA-Med \cite{li2023llava}, which demonstrates significant performance degradation on out-of-distribution data, XrayClaw maintains a stable lead even when compared to general-purpose models such as GPT-4o \cite{hurst2024gpt}. Although MedRAX \cite{fallahpour2025medrax} achieves a slightly higher score on Rad-Restruct, XrayClaw remains highly competitive with a second-best accuracy of 66.3\%. This consistent performance across diverse clinical benchmarks suggests that the iterative refinement through ComPO enables the model to capture invariant clinical features rather than overfitting to specific dataset characteristics.

Furthermore, the results on CheXbench demonstrate the zero-shot transferability of the diagnostic reasoning process. By decomposing the interpretation task into specialized agent roles and enforcing an adversarial audit, XrayClaw effectively mitigates the impact of domain-specific biases. The overall performance gain of 2.6\% over the strongest multi-agent baseline, MedRAX, confirms that the integration of analytical cooperation and holistic competition provides a robust foundation for reliable domain generalization. Consequently, these findings validate the potential of XrayClaw for practical deployment in diverse clinical environments where data distributions may vary significantly from the training domain.

\section{Conclusion}

In this paper, we presented XrayClaw, a novel multi-agent framework designed to enhance the trustworthiness and interpretability of Chest X-ray diagnosis. By orchestrating a cooperative-competitive architecture, the framework effectively reconciles the granular evidence extraction of a multi-agent pipeline with the holistic auditing of a senior attending radiologist. Central to this approach is the Competitive Preference Optimization mechanism, which penalizes logical hallucinations by enforcing mutual verification between distinct diagnostic pathways. Extensive empirical evaluations across the MS-CXR-T, MIMIC-CXR, and CheXbench benchmarks demonstrate that XrayClaw achieves state-of-the-art performance in diagnostic accuracy, reasoning fidelity, and zero-shot domain generalization. These results substantiate that the integration of analytical cooperation and competitive auditing establishes a robust foundation for reliable medical imaging analysis. Ultimately, XrayClaw provides a scalable and interpretable paradigm for automated CXR interpretation, validating its potential for practical deployment in diverse and complex clinical environments.

%
%
%
\bibliographystyle{splncs04}
\bibliography{ref}

\end{document}